\documentclass{article}

\usepackage{arxiv}

\usepackage[utf8]{inputenc} 
\usepackage[T1]{fontenc}    
\usepackage{hyperref}       
\usepackage{url}            
\usepackage{booktabs}       
\usepackage{amsfonts}       
\usepackage{nicefrac}       
\usepackage{microtype}      
\usepackage{lipsum}
\usepackage{graphicx}
\graphicspath{ {./images/} }

\title{A Storytelling Robot managing Persuasive and Ethical Stances via ACT-R: an Exploratory Study. 
\\}

\author{
  Agnese Augello\\
  ICAR-CNR, Italy\\
  Palermo \\
  \texttt{agnese.augello@cnr.it} \\
   \And
  Giuseppe Città\\
  ITD-CNR, Italy\\
   Palermo \\
  \texttt{giuseppe.citta@cnr.it} \\
  \And
 Manuel Gentile\\
  Università di Torino, Dipartimento di Informatica, Italy\\
   ITD-CNR, Italy\\
  Torino \\
  \texttt{manuel.gentile@unito.it} \\
   \And
 Antonio Lieto\\
  Università di Torino, Dipartimento di Informatica, Italy\\
   ICAR-CNR, Italy\\
  Torino \\
  \texttt{antonio.lieto@unito.it} 
}

\usepackage{natbib}
\usepackage{amssymb}
\usepackage{longtable}
\usepackage{graphicx}
\usepackage{hyperref}
\usepackage{listings}

\usepackage{times}
\usepackage{url}
\usepackage{xcolor}
\usepackage{soul}
\usepackage[utf8]{inputenc}
\usepackage[small]{caption}
\usepackage{amssymb}
\usepackage{algorithm}
\usepackage[noend]{algpseudocode}
\usepackage{multirow}
\usepackage{subcaption}
\usepackage{tabularx}
\usepackage{booktabs}
\usepackage[flushleft]{threeparttable}
\usepackage{caption}
\usepackage{float}

\begin{document}
\maketitle







\begin{abstract}

We present a storytelling robot, controlled via the ACT-R cognitive architecture,  able to adopt different persuasive techniques and ethical stances while conversing about some topics concerning COVID-19.  
The main contribution of the paper consists in the proposal of a \emph{needs-driven} model that guides and evaluates, during the dialogue, the use (if any) of persuasive techniques available in the agent procedural memory.
The portfolio of persuasive techniques tested in such a model ranges from the use of storytelling, to framing techniques and rhetorical-based arguments. To the best of our knowledge, this represents the first attempt of building a persuasive agent able to integrate a mix of explicitly grounded cognitive assumptions about dialogue management, storytelling and persuasive techniques as well as ethical attitudes. 
The paper presents the results of an exploratory evaluation of the system on 63 participants.


NOTE: \emph{Paper to appear in the International Journal of Social Robotics (Springer)}.
\end{abstract}






\section{Introduction}
\label{intro}
In the last decades, the field of Human-Computer Interaction (HCI) has started to focus its attention on the design and implementation of artificial systems “orienting”  attitudes and/or behaviours of a user according to a predefined direction. This growing sub-field, studying the so-called Persuasive Technologies, concerns a variety of system typologies that can adopt different strategies to pursue their goals. Building persuasive robots able to interact with human beings on a specific topic (or in a  multi-domain setting)  in a realistic and persuasive way, represents also an open problem and research challenge in Social Robotics.
To this aim, a strategy often used in human-human communication to make people reconsider their behaviour and beliefs, and similarly proposed in human-robot interaction, is to exploit storytelling to let people identify themselves with the characters or roles in a story in order to understand different perspectives and needs. In the design of a persuasive system, in addition, it is also important to not ignore the ethical dimension: i.e. an intelligent artificial system should be able to make decision and act in an ethical way, taking into account norms of social practices  and needs of other individuals. 
In this paper, we present the model of a persuasive robot relying on a plurality of cognitive assumptions based on dialogue modelling, storytelling, ethical perspectives, as well as a selection of rhetorical techniques studied in the field of cognitive science, logic and argumentation theory\footnote{The overall model and architecture presented in this work is entirely equipped for being used in a real robotic agent. However, due to the COVID-19 restrictions concerning  access to our lab and the impossibility of hosting external people for experiments, the evaluation phase has been conducted on a simulated robot in a 3D environment.}.
The overall assumptions of the model are controlled and organized via the ACT-R cognitive architecture. In particular, the system is able to adopt different kinds of persuasive techniques and ethical stances in order to carry out a  conversation on COVID-19 rules and vaccines. The system investigates the awareness (and knowledge) of the interlocutor about the COVID-19 rules and her/his willingness to get vaccinated against the disease.
The system, in addition, exploits a storytelling strategy to emphasize the importance of considering different situations in which a user could find herself/himself when asked to follow the COVID-19 rules (e.g. one could be allergic to the material of a mask).
Finally, the system can also manage the entire dialogue by following (or not) an ethical setting: i.e. by taking into account some particular conditions of the user (like the allergic case just mentioned).

To the best of our knowledge, the proposed model represents a first attempt to integrate all the above-mentioned aspects in a cognitive architecture that is usable both in a robotic setting and in virtual environments.

The rest of the paper is organized as follows. In section \ref{related} we introduce the related works concerning the different dimensions integrated into the proposed model, namely: the persuasive one, the one concerning  storytelling abilities in robots and, finally, the one concerning the modelling of robotic ethical behaviour. We  position our work along such dimensions by providing an overview of the related work outlining our approach.
In sections \ref{cogmod}, we  present the agent model integrated into the ACT-R cognitive architecture. In section \ref{impldetails} we provide further implementation details and an example of interaction, while in section \ref{results} we describe the results of the evaluation carried out. Finally, in \ref{conclusion} we discuss the obtained results and envision  future works to be carried out.

\section{Related Works}
\label{related}

\subsection{Persuasive interaction\label{Persuasive interaction}}
Persuasive technologies can adopt several strategies to change the attitudes and behaviours of their target users.
In this work, we ground our persuasive interaction on two different but interconnected theories coming from cognitive psychology, namely the ELM theory (Elaboration Likelihood Model (ELM) elaborated by [\cite{petty1986elaboration}]) and the dual process theory of reasoning elaborated by Kahneman [\cite{kahneman2011thinking}]. 
According to the ELM theory, there are two different information processing paths (or routes) followed by a message: one, the peripheral route, where the processing is based on scarce attention and on the focus on surface elements (and as such more akin to trigger fast and automatic cognitive mechanisms that are not subject to any form of deliberative control) and another one, the central route, through which the information is processed in a more deliberative, controlled and logical way.
A similar distinction is made in the dual process theories of reasoning, suggesting that our decision making processes are governed by two types of interacting cognitive systems, which are called respectively system(s) 1 and system(s) 2. Systems of the type 1, referred also as S1, operate with rapid, automatic, associative processes of reasoning. They are phylogenetically older and execute processes in a parallel and fast way. Type 2 systems, referred also as S2, are, on the other hand, phylogenetically more recent and are based on conscious, controlled, sequential processes (also called type 2 processes) and on logic based rule following. As a consequence, if compared to system 1, system 2 processes are slower and cognitively more demanding. By following such theories, our working hypothesis is that persuasive strategies should trigger heuristic-driven and fast (e.g. type 1) processes elaborated via the the ELM peripheral route. To this end, we selected persuasive strategies based on some well known rhetorical arguments and  the framing technique. Such persuasive strategies, indeed, exploit the peripheral route [\cite{petty2009elaboration}],[\cite{petty2011elaboration}],[\cite{o2013elaboration}] and, as such, are assumed to be processed automatically (thus eluding some deliberative forms of cognitive control that are executed, on the other hand, in the central route of information elaboration). In particular, by following [\cite{lieto2013unveiling,DBLP:journals/psychnology/LietoV14}], in the present work we have considered rhetorical arguments based on inferential schemas that, even informally invalid, appear as plausible and therefore are psychologically persuasive [\cite{cohen1993introduction,hamblin1970fallacies}]. Such kinds of arguments should not be considered irrational. Indeed, in the context of ecological approaches to rationality and cognition, it has been pointed out that they can have a proper heuristic value [\cite{walton1998new}]. 

Recently, the exploitation of this kind of cognitive tendencies to design and evaluate the effect of nudging elements able to guide people choices in digital environments is gaining widespread attention in Human-Computer Interaction [\cite{Theocharous:2019:PHC:3314183.3323453}]. This sub-field is known as Digital Nudging [\cite{schneider2018digital}]. Our work can be ascribed to this class of analysis and, in the rest of this section, we present the persuasive arguments exploited for the design and implementation of the proposed system.

The first adopted argument in our model is known  as ``appeal to the majority'' (or \textit{Argumentum Ad Populum}). It consists~of accepting a certain thesis based on the mere fact that the majority of people accept it. Its typical characterization is the following: ``Most people think that X is true/false, then X is true/false'' (where ``X'' can be any statement).
This argument can be compared to those strategies commonly used in the realm of persuasive technologies, which owe their persuasive potential to the exploitation of social dynamics. In particular, Fogg refers to well-known social psychology theories (e.g.,~social~comparison and conformity [\cite{turner1991social}]), which can be extended to include computer technologies. According to social comparison theory, people who are uncertain about the way they should behave in a situation proactively collect information about others and use it to build their own attitudes and behaviours.
By contrast, conformity theory focuses on normative influence, stating that people who belong to a group usually  experience a pressure to conform to the expectations of the other group members. 
Another argumentation schema used in our work (and, in general, in both digital artifacts  and human-human interaction) concerns the perceived credibility [\cite{DBLP:conf/chi/FoggMLOVFPRSST01}]. Such credibility is known to be affected by the so-called halo effect [\cite{Dion72whatis}], according to which a positive evaluation on a specific aspect (e.g., physical attractiveness) produces a halo which determines an extension of such an evaluation to other, unrelated, aspects (e.g., expertise in a certain field). This aspect is crucial in the so-called  ``appeal to the authority'' (also \textit{Argumentum Ad Verecundiam}). 
Such argument refers to cases of inappropriate transfer where some theses are assumed to merely hold because the people asserting them are, wrongly, assumed to be authorities about a certain topic due to their achievements and fame obtained in other, unrelated fields.
Finally, another well-known persuasive technique adopted in our work, that is based on a well-known cognitive tendency in human decision making, is the so-called  \textit{framing} effect [\cite{lakoff2014all}], [\cite{tversky1981framing}]. It refers to the role of the context in shaping people's decisions. In fact, using a particular wording instead of another might determine a different configuration of a given problem that consequently, may~lead to a given interpretation of a sentence's meaning. A classic example of framing for providing information about food nutrition is the following: ``The food X contains 60\% of lean meat. Therefore,~X~is sustainable''. Of course, this interpretation is misleading since the conclusion cannot be drawn from the premises, but it is framed by them (e.g., the reverse of the frame, in fact, is that The food X contains 40\% of fat meat).
Another well-known corollary of the framing effect consists of the fact that there is an asymmetry between prospective losses or wins in the  people choice's architecture [\cite{tversky1981framing}]. This effect is known and theorized in the \emph{prospect theory} and can be roughly explained as follows: people prefer prospective choices that lead them not to lose something instead of choices that could provide them with the possibility of winning something else. This means that framing a given context as a possible loss (negative framing) should be a more sensitive and persuasive move to induce people towards a given behaviour. The loss aversion predicted by the prospect theory is related to another well-known effect: the scarcity one.  In this setting, this means that the more something is perceived as scarce, the more the prospective loss is valued as problematic (and this usually leads to a less risk-seeking behaviour [\cite{shimizu2007prospect}], [\cite{palmer1999decoupling}] or, in our case, to an action aimed at removing this sense of potential loss). All these three above mentioned techniques are adopted in our interaction model.
 

\subsection{Ethical models of persuasive behaviour for robotics}
\label{virtue}

The necessity of robots with ethical capacities is starting to become widely recognised (e.g., [\cite{deng2015machine}], [\cite{arkin2011moral}], [\cite{moor2011nature}], [\cite{dignum2018design}]). Nevertheless, only a few studies have implemented models of roboethics. So far, most work has been either entirely theoretical (e.g., [\cite{mackworth2011architectures}], [\cite{wallach2008moral}] or based on simulation (e.g. [\cite{arkin2011moral}]) and, overall, the existing developed architectures for ethical robots have been entirely based on logic frameworks ([\cite{arkin2011moral}], [\cite{bringsjord2006toward}]) or rooted in specific cognitive theories [\cite{vanderelst2018architecture}]. 

The approach followed in this work, on the other hand, is the first one built within the constraints of a well-established unifying cognitive architecture like ACT-R [\cite{anderson2004integrated}]. Such architecture, in fact, exploits a variety of integrated mechanisms for the emergence and simulation of human-like  behaviour in artificial cognitive  systems.

Within the context of the implemented persuasive dialogue practice, the deepening of an ethical dimension is another innovative aspect of the work carried out. In the context of social robotics, in fact, both reflections related to the ethical dimension and reflection related to persuasion have received a great deal of attention in the literature marching in parallel and sometimes intertwining [\cite{kim2021robots}], [\cite{Zhu2020}]. However, our approach takes into account a particular ethical stance (that will be introduced below) related to the particular dialogical condition that we have tested. 
In general, among the various ethical theories in the HRI field, two have been mainly successful: the consequentialist approach [\cite{10.2307/3749051}], and the deontological approach [\cite{alexander2007deontological}]. They have been referred to as action ethics in that they have focused on robot performance as managed by the robot's ability to make correct decisions based on sound moral principles. On the one hand, consequentialist ethics refuse to define a priori moral values and argue that an action should be evaluated by relating it to its effects, whereby behaviour is right if it produces good consequences [\cite{10.2307/3749051}]. According to this perspective, an ethical decision by a robot that “directs" action in an ethical manner finds its justification in what humans find beneficial. The core of this ethical perspective is the consequence; whether an action is right or wrong depends on the consequences it has [\cite{7736246}]. A representative case of consequentialism is Utilitarianism [\cite{mill18611998}]. According to it, an action is best if it maximizes the general welfare. The general principle can be synthesized in the statement “the greatest happiness for the greatest number" [\cite{jeremy2015introduction}]. 
On the other hand, deontological ethics judge the morality of a choice or an action on criteria that have nothing to do with the states of affairs that those same choices/actions produce. According to a deontological perspective, some choices cannot be justified by the effects they produce, even if they are morally good. Duties and rights are at the heart of this ethics. The action of telling the truth, for example, is good because one has to do so [\cite{7736246}].
These theories have been successful in HRI because they both allow one to think of the morality of a choice or action as manageable by a set of principles or rules that are within reach of a machine.
According to Cappuccio [\cite{cappuccio2020can}] each theory that can be traced back to one of these two approaches requires robots that acts ethically. The ethicality of such actions, according to these perspectives, would lie in the fact that the robots are sufficiently intelligent to correctly apply to real-life circumstances specific top-down decision-making processes inspired by general ethical principles.
However, a third ethical path seems to be emerging within HRI: Virtue Ethics [\cite{coeckelbergh2021use}], [\cite{cappuccio2020can}]. Unlike action-based consequentialist and deontological ethical theories, virtue ethics as agent-based ethics focuses on the ethical conduct of an agent in terms of the realization of the positive dispositions embedded in the agent's nature. This ethical model has its roots in Aristotle's ethical perspective [\cite{Aristoteles2005En/A}] according to which virtues are primary qualities of persons and their lives [\cite{crisp1996should}]. Virtue Ethics is defined as agent-based ethics. It suggests that such an ethical model focuses not on the moral value of the actions performed by the agent, but on those traits of his/her character enabling him/her to be virtuous. However, it is necessary to stress that such a perspective does not exclude actions from the ethical dimension but focuses on the performer, on his/her practice (habitual action) in terms of performance which can be executed in different ways.  According to Gascón [\cite{gascon2016virtue}], the question of the Virtue Ethics theories is not “What should I do in this situation?" but “What kind of person should I be in this situation?". 
Such models propose Virtue Ethics as an ethic of practical behaviour implemented within a series of acts, namely through habit [\cite{Aristoteles2005En/A}]. “Habit" (or “\textit{habitus}") translates, albeit incompletely, the Greek term “\textit{hexis}" and emphasizes precisely the fact that habits (states, ways of being) are created, manifested, consolidated, and learned in the context of the repetition of a pattern of actions [\cite{bourdieu1980sens}]. They generate expectations and shape cognitive and emotional states [\cite{coeckelbergh2021use}]. 
The Virtue Ethics opens a focus on the dispositions (\textit{hexeis}), praiseworthy or deplorable, that take the name of “virtues" or “vices". The dispositions can be conceived as habitual models of behaviour to be implemented in daily practice and whose principles are not based on abstract systems of rules [\cite{cappuccio2020can}].
Among everyday practices, persuasion has been the subject of in-depth reflection on the ethical dimension. Specifically, the approach of the ethical dimension to persuasive practices dates back to Aristotle who, in his investigation on “what can be persuasive" [\cite{Aristoteles2010R/A;}], outlines those typical traits that an arguer should possess to increase the perception in his/her interlocutor that he/she is a trustworthy and, therefore, convincing person [\cite{piazza2004linguaggio}]. In this context, the “honesty" of the arguer is one of the key characteristics that should be brought into play in making a convincing argument. The honest arguer, having a deep “sense of circumstances", adapts to situations flexibly. She/He is helpful and sympathetic [\cite{piazza2008retorica}]. She/He is guided by her/his wisdom (practical intelligence), her/his benevolence towards the interlocutor and her/his moral virtue [\cite{Aristoteles2005En/A}].
Today, the Aristotelian perspective is borrowed from Virtue Argumentation Theory (VAT) [\cite{Aberdein2010}]. It proposes an approach to the virtue of argumentation that focuses on the arguer, his/her attitudes, and behaviour [\cite{gascon2016virtue}]. According to Gascón [\cite{Gascon2018}], two categories of virtues can be identified: reliabilist virtues and responsibilist virtues. The formers have to do with the skills of the arguer. The latter virtues have to do with the arguer's attitude, character, behaviour, and habits. According to this perspective, an arguer should not only present convincing arguments but should be open to different points of view, willing to put his or her beliefs to rational criticism, and be respectful of other points of view [\cite{Gascon2018}]. These traits, within this perspective, are outlined in a system of argumentative virtues [\cite{Aberdein2010}]. It offers a conceptual framework through which to study argumentation as a genuine social practice rather than a static product of rules. Within this ethical system, in line with the Aristotelian viewpoint, open-mindedness emerges as the “critical virtue" of the arguer [\cite{kwong2016open}]. In particular, it is defined as “the ability to listen carefully, the willingness to take seriously what others say and, if requested, the determination to adopt it as one's own" [\cite{cohen2009keeping}].  
Concerning the practice of persuasion in the context of human-robot interaction, the attempts to  “regulate'' through an ethical model the persuasive practice of a robot are sporadic [\cite{ghazali2020persuasive}], [\cite{zhu2020blame}]. Moreover, to the best of our knowledge, there are no studies that deal with implementing an ethical persuasive practice of a robot according to instances referable to the Aristotelian model, the VAT perspective, or Virtue Ethics in general. This is, on the other hand, the approach we explicitly have taken in this work.



\subsection{Persuasive Storytelling Robots}
\label{narrative}

Storytelling has always been an important component of human communication that is also exploited with persuasive purposes to support individuals in understanding different perspectives, stimulating insight and behaviour change through different narrative strategies [\cite{divinyi1995storytelling}].  Narrative strategies, indeed, enabling an identification with  the content of the stories and establishing an emotional connection with the characters, can lead people in reconsidering their own beliefs and behaviours [\cite{Zacks}], [\cite{koivula2020using}] or  obtaining more positive responses with respect to more direct approaches like behavioural instructions [\cite{faddoul2020quantitative}].

Artificial intelligence is often used to manage the narration and to model the characters of a story as autonomous agents. In some cases, the users can interact with the characters of the story and give them advices, influencing the evolution of the narration, with the result of a  better comprehension of story dynamics [\cite{aylett2005fearnot}], [\cite{bono2020social}], [\cite{evans2013versu}].
Since storytelling is an embodied activity deeply based on the use of gestures and postures,  storytellers robots are often proposed instead of virtual agents, most commonly for educational purposes with the result of a greater engagement [\cite{kory2014storytelling}], [\cite{striepe2017there}], [\cite{park2019model}], [\cite{conti2020robot}], [\cite{augello2018introducing}].
Storytelling robots have been also used to raise and foster awareness about sensitive themes  [\cite{bono2020social}], and to influence behaviour [\cite{Paradeda2017}], [\cite{paradeda2020persuasion}]. In particular, starting from the assumption that the success of a persuasion strategy depends on a reference target group, Paradeda et al. implemented in a robot a   persuasion model based on a categorization of humans personality traits, with the aim to  influence the listener in choosing the actions to perform according to a story plot  [\cite{Paradeda2017}],  [\cite{paradeda2020persuasion}]. As an example, they tested such a personalized strategy in a scenario where the robot persuaded participants to make monetary donations [\cite{fides}]. 

In this context, the effect of the use of social cues by the robot has been analyzed, considering that people will be more socially responsive to the agent that has more social cues. In [\cite{ham2015combining}] it was verified that  robots can become more persuasive when they look at the person to persuade and that this effect is stronger when they use gestures, but only when looking at the person, as it also happens in interaction between humans.
However, as in the case of human-human interaction, strong persuasive attempts and a forceful language could lead to a negative outcome and to the so-called psychological reactance, i.e., people could not accept the advise or even could behave in the opposite way [\cite{ghazali2018influence}].
Therefore, it is important to pay attention to these issues when designing a persuasive agent so as to have communication that is as persuasive as possible and avoids reactance.
The wizard of oz study described in [\cite{ghazali2018influence}] analyzed the impact of social cues of artificial agents on reactance, considering three different levels: 1) low social agency: absence of a robot and textual prompts on a screen; 2) medium social agency: a robot with a human-like face and minimal social cues; 3) high social agency: robot using multiple social cues. The latter caused greater reactance because of the strong affective tone of voice and the excessive pressure to change participants' choices. Moreover, the medium social agency caused lower reactance in comparison with the low social agency because it has been  perceived as a less forceful way of giving advice, compared to text.

In our work, we propose the use of a  storytelling-based model embedded in a cognitive architecture that can be used in a robotic platform. Such a model uses the narrative and persuasive strategies exposed in section \ref{Persuasive interaction} to, through well-defined phases, lead the listener to have greater knowledge and awareness of the risks involved in not respecting the  COVID-19 rules.
In agreement with what has been previously discussed, we have enclosed the persuasive strategy within a narrative context instead of directly provide behavioural indications.

In particular: the conversational practice is managed by following a typical narrative arc, starting from an initial phase where the agent introduces the main elements of the discussion topic, and at a given point reaching a climax where a conflict must be resolved. The climax is generated voluntarily by the agent to let the user experience the situation of someone who, because of some problem, cannot strictly follow the COVID-19 rules. This situation of conflict is forced to further emphasize the importance of following the rules to protect most vulnerable individuals.  Furthermore, an open-mindedness ethical attitude displayed by the agent causes him to suggest some possible ways out to resolve this conflicting situation by showing alternatives to preserve the health of the individual. 
Since the narrative is interactive, the agent will deepen and argument the topic according to the answers given by the user to questions aimed at assessing her/his level of knowledge, her/his intention to pursue the anti COVID-19 rules as well as her/his intention to be vaccinated against COVID-19.

\section {Agent’s Cognitive Model}

\label{cogmod}

In this section, we discuss  the concepts at the basis of the formalization in ACT-R of the agent’s cognitive model. For the sake of self-containedness, we also briefly introduce the main features of ACT-R. 

ACT-R [\cite{anderson2004integrated}] is  a  cognitive  architecture  explicitly  inspired by theories and experimental results coming from human cognition.  Here the cognitive mechanisms concerning the knowledge level emerge from the interaction of two types of knowledge:  declarative knowledge, which encodes explicit facts that the system knows, and procedural knowledge, which encodes rules for  processing declarative knowledge.  In particular, the declarative module is used to store and retrieve pieces of information (called chunks, composed of a type and a set of attribute-value pairs, similar to frame slots) in the declarative memory. ACT-R employs a sub-symbolic activation of symbolic conceptual chunks representing the encoded knowledge.  Finally, the central production system connects these modules by using a set of IF-THEN production rules. 

The symbolic component of ACT-R includes a set of modules, where each of them is associated with a specific \textit{buffer} that serves as the interface with that module.  The ACT-R activities rely on the coordinated work of these modules with their buffers.
For example, an  \textbf{audio} module is exploited by the agent of the ability to perceive sounds, by interfacing with the \textbf{\textit{aural-location}} buffer managing requests about the source of the sound  and a \textit{\textbf{aural buffer}}, keeping track of what has been heard.  A \textbf{speech} module provides the agent of the ability  to communicate with other cognitive models using the buffer \textit{\textbf{vocal}}. 
The sub-symbolic component consists of a series of parallel processes responsible for the learning processes in ACT-R.

For the purposes of this work, we decided to adopt this architecture as an integrated blueprint for the development of our model (rather than as a tool in which to test a simulative account of human cognition [\cite{lieto2021cognitive}]),  since its overall knowledge processing mechanisms -  based on the continuous interaction between long and short term memories and on the activation of relevant pieces of knowledge obtained via a spreading mechanism [\cite{lieto2018knowledge}]- are the ones more compliant with the requirements of our model. 

In particular, the use of the ACT-R spreading activation mechanisms for the retrieval and activation of the rules governing the dialogue management (driven by both a narrative structure and by an internal model of needs-action cycle) represented, from our perspective, an important aspect for developing a non-deterministic behaviour, emerging from the devised model.

The equation for the activation of a chunk $A_i$including spreading activation is defined as:


\[ A_i = B_i + \sum_{j} W_j S_{ji} + e \]

where $B_i$ represents the base-level activation of a chunk,  reflecting the recency and frequency of the chunk activation;  of the chunk activation; the elements j being summed represent the chunks which are in the slots of the goal chunk; the weighting $W_j$  is the amount of activation from source j; $S_{ji}$ represents the strengths of association from source j to chunk i and e is a stochastic noise value, as described in [\cite{anderson2004integrated}].
 
%
%

Such mechanisms allow, in principle, the design of a flexible decision making strategy that we have employed in our agent. In addition, the overall mechanisms of ACT-R allowed us to ground and constrain our model on the information processing mechanisms of the architecture that - in our case - despite not being directly used for measuring the compliance with human performances, allowed us to reuse a framework already developed for integrating intelligent abilities and modules in a cognitively well founded way. This choice provided the agent with the ability of autonomously managing its decision making according to a non sequential narrative flow.

In particular, as  mentioned, the implemented ACT-R cognitive model has been defined to enable the agent to dynamically manage the dialogue practices dealing with narrative and persuasive strategies about the controversial topic of COVID-19. More specifically, in this work, the aim of the conversational practice was to evaluate to what extent the combination of persuasive and narrative strategies, in different ethical settings, can contribute to the acceptance of the respect of COVID-19 rules and the willingness to being vaccinated.

Our model is loosely inspired by the planned behaviour theory [\cite{ajzen1985intentions}] suggesting  the agent must conduct the conversation by collecting information about the user's beliefs and attitudes in order to estimate the user’s intention and  therefore evaluate the persuasive strategy to use. Then, when necessary, it must use argumentative examples to change the intention held by the listener or reinforce it.

\begin{figure}[h!]
\begin{center}
\includegraphics[width=1.08\textwidth]{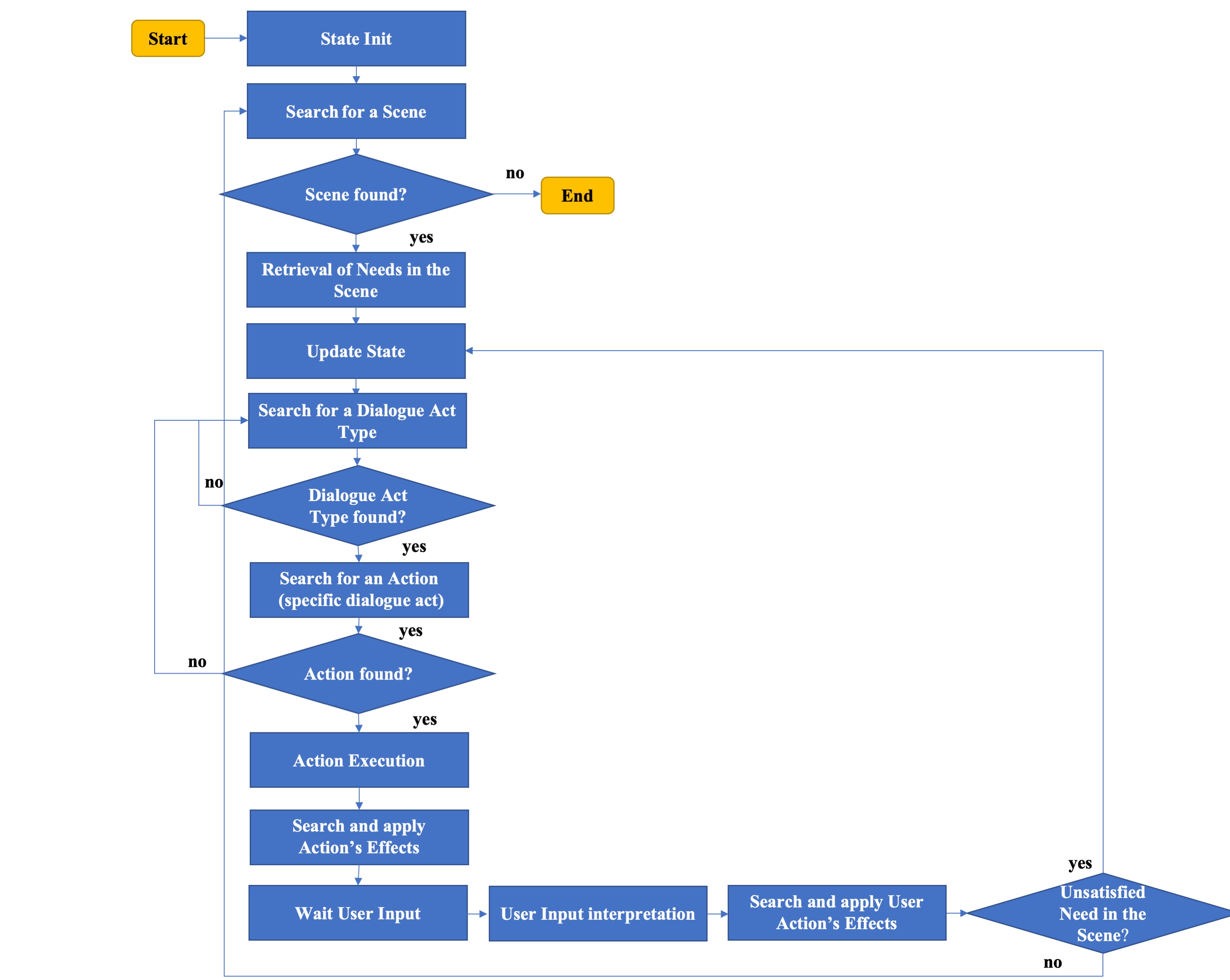}
\end{center}
\caption{Flow of the agent's reasoning process}
\label{flow}
\end{figure}

In our model, the agent pursues the fulfilment of his goal by planning a sequence of scenes related to its specific needs and carried on by selecting dialogue acts. These conversational choices are performed by the agent according to an Information State approach [\cite{Traum2003}], by evaluating and updating from time to time, information about the participants of the practice and the state of the dialogue. In particular, the Information State keep track of the following elements:
\begin{itemize}
\item the level of the user's knowledge and her/his intention about the main topic of the conversation (COVID-19), and subtopics (such as COVID-19 rules, vaccines), inferred by the answers provided during the conversation.
\item the current needs of the agent (see the section below); 
\item the current scene (that in our implementation identifies a specific topic of conversation or a phase of the conversational practice) 
\item the previous scene 
\end{itemize}
The information state is updated according to the actions performed by the participants and a set of updating rules by taking into account the effects caused by the actions of the interlocutors.
The behaviour of the agent is therefore influenced by the context of the conversation, and by its  needs that emerge according to the information state. This allows also balancing both conversational norms and agent's personality. The flow of the described reasoning process of the agent is detailed in  Figure \ref{flow}.

\begin{figure}[h!]
\begin{center}
\includegraphics[width=1\textwidth]{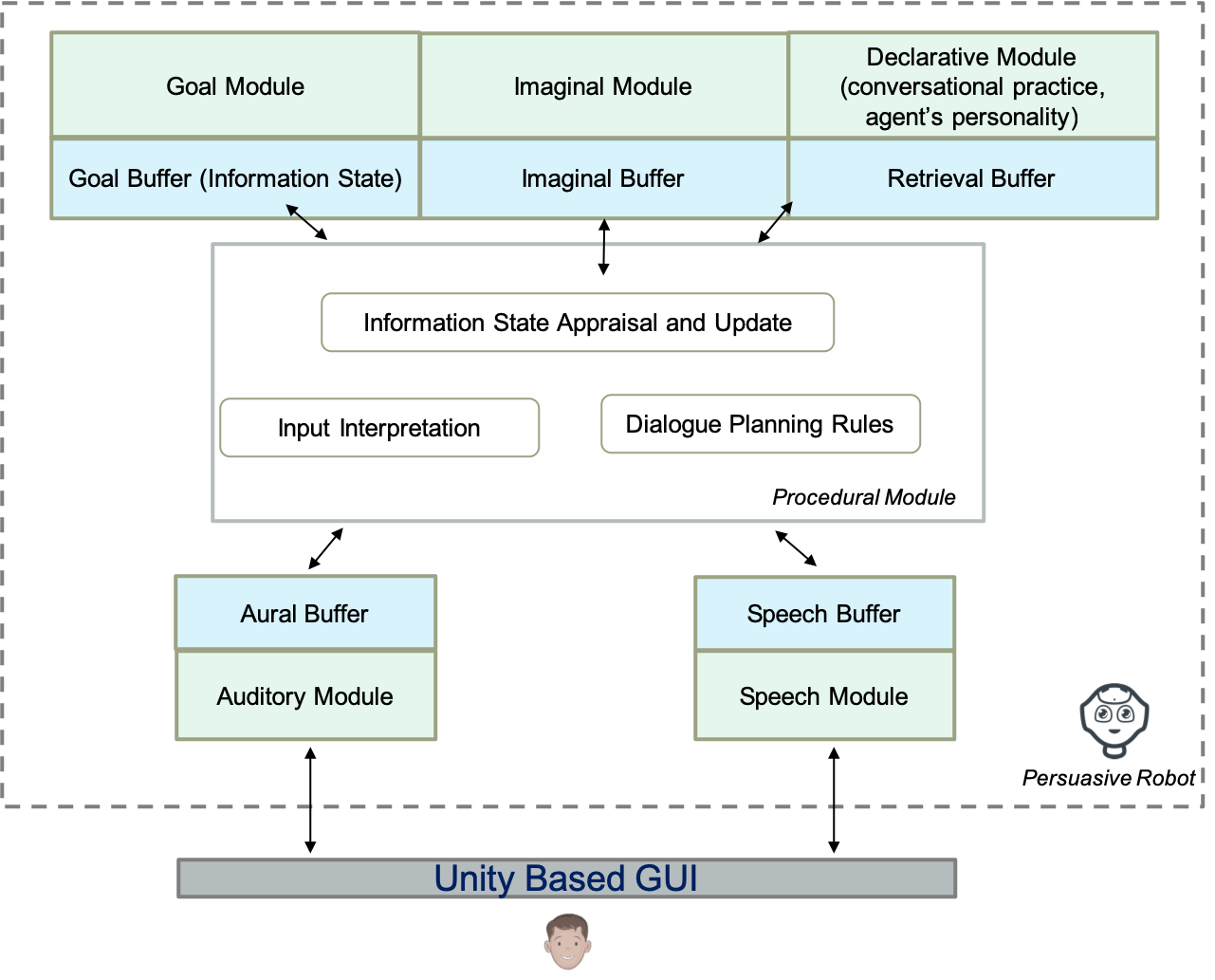}
\end{center}
\caption{The ACT-R based Architecture and its use in the implemented model.}
\label{arch}
\end{figure}

The Figure \ref{arch}, on the other hand, shows the modules of ACT-R exploited by the agent 
to manage the conversational practice. 
\textit{Aural}  and \textit{Speech} modules and their buffers allow the agent to respectively analyse the input and produce an output through a GUI interface. The knowledge is composed of a \textit{declarative} and \textit{procedural} part. The declarative memory handles the creation and archiving of the facts (\textit{chunks}), that represent the atomic knowledge of the model through lists of (key, value) pairs. The declarative module manages the knowledge about the conversational practice and what characterizes the personality of the character (its ethical profile and the formalization of its needs). The imaginal module and its buffer are used by the agent to manage all the elaborations and it is used as a short term memory. The procedural module processes and interprets the user input and it is responsible for the planning of the conversation, through the processing of rules, that take into account the Information State. It is also responsible for the Information State update.

In the next sections we detail the main elements characterizing such Information State-based approach: Needs and Dialogue Acts.


\subsection {Needs}
\label{needs}

By following the discussion about motivated cognition of [\cite{bach2009principles}], our  agent is characterized by a motivational component implemented as a set of needs, that, when unsatisfied, drive the dialogue acts of the agent,  aimed at compensating the difference between expected and current value.

Different phases of the dialogue (scenes) lead to the emergence of agent's needs, that  can be fulfilled by actuating different dialogue acts.  

In particular, we have outlined four categories of needs: social, cognitive,  argumentative and narrative. 
Social needs are introduced to model a demand of the agent to socialize and be compliant to social obligations. In this category we modelled a \textit{social affiliation need}, that is unsatisfied at the beginning of each dialogue, since when the agent meets a new user, it has the need to start the conversation with a greeting and a self introduction. 
When the \textit{social affiliation need} is satisfied, then   cognitive needs emerge since it is necessary for the agent to acquire information about the user. In particular, in the analyzed scenario, it is important for the agent to be aware of the knowledge of the user (\textit{competence need}) about the COVID-19 topic, and his intention to follow the COVID-19 rules (\textit{intentional\_assessment}).
When the agent recognizes that the intention of the user is low, an \textit{argumentation need} emerges,  leading to the accomplishment of persuasive arguments  to increase such a value. 
Moreover, to keep the involvement of the user high, the agent has a need to follow a narrative structure to let the user experience a climax-based situation (\textit{climax need}), as detailed in the following section \ref{da}.

Finally, another need assumed in our model concerns the ethical virtue of the agent with respect to its  dialogue attitudes. By following the Virtue Ethics and VAT theory frameworks discussed in \ref{virtue} section, this need has been called \textit{open mindedness}  since it refers to the willingness of the agent to consider the user perspective (including ideas and opinions different from its own) during the dialogue. As a consequence, in our model, it is an indicator of the “ethical willingness" of the agent. This need, differently from the other ones, can emerge during the interaction only if the agent is set, at the beginning of the interaction, with this “ethical profile" and eventually emerges only in a conflicting situation.

Below we detail the dialogue acts that can be activated to fulfil the aforementioned needs.


\subsection {Dialogue Acts }
\label{da}

The dialogue acts, i.e. the communicative functions associated to each action in the dialogue, have been defined and named by considering the ISO 24617-2 standard for dialogue annotation [\cite{bunt2019guidelines}]. Such acts are directly activated by the needs discussed in the previous section. 
For example, the \textit{argumentation need} activates one of the persuasive arguments presented in section \ref{Persuasive interaction} and, therefore, one of the actual dialog acts in which such arguments are employed.  Similarly, there are acts having a “\textit{social obligations management}” function, associated to specific scenes, and implemented to respond to the  \textit{social affiliation need} of the agent (i.e. the need of the agent to socialize and be compliant to social obligations). For example, it is socially correct for the  agent to start the first scene of the dialogue with an introduction act having both the communicative functions of “Initial Greeting”  and “Initial Self-Introduction”, to make himself known to the interlocutor. The agent also knows that at the end of the practice it must perform an action having an “Goodbye” communicative function.

Other acts have  general-purpose functions, such as making certain information available to the listener (“\textit{informative acts}") or obtaining certain information from her/him (“\textit{question acts}").
Concerning the “\textit{informative acts}"we considered the following fine grained classification:
\begin{itemize}
    \item \textit{inform}: used by the agent to provide information about a topic $t$, when the knowledge of the user about the topic is low;
    \item \textit{reinforce}: used by the agent to provide information about a topic $t$, when the knowledge of the user about the topic is medium;
    \item \textit{argument}: used by the agent to argument when the intention of the user toward a specific topic $t$ is low;
\end{itemize}

In particular, the \textit{argument} dialogue acts, implemented to satisfy \textit{the argumentation need}, are those equipped with the persuasive techniques described in \ref{Persuasive interaction} (e.g. ad verecundiam, framing, or ad populum; the adoption of one of these techniques is randomly selected).

As anticipated in section \ref{needs}, also a \textit{climax} is introduced by the agent itself to expose the listener to possible alternatives in a critical situation. 
This is accomplished through a role-playing strategy, where the player is invited to assume the identity of an individual who is affected by a particular condition  that makes it infeasible to respect some rules.

Therefore, there are also  \emph{exception}  and  \emph{substitution} actions designed to fulfil the \textit{climax need} and used by the agent to respectively  introduce and manage such conflicting situations. 



The former is used by the agent to introduce a role for the listener that forces an exception for a previous agent's argumentation ($arg1$). This action  introduces a condition $cond$ indicating a circumstance in which the argument specified by $arg1$ does not hold. 

The latter is used to deal with the conflicting situation  suggesting a substitution of the argumentation $arg1$ with an alternative $arg2$ that is compatible with the user condition $cond$.

This alternative, offering a possible way out to the canonical argument (and therefore considering the user condition), is only considered if the agent has an open-mindness “ethical profile" (in the sense defined above). 




\section{Implementation details and an example of interaction}
\label{impldetails}

\begin{figure}[h!]
\begin{center}
\includegraphics[width=0.8\textwidth]{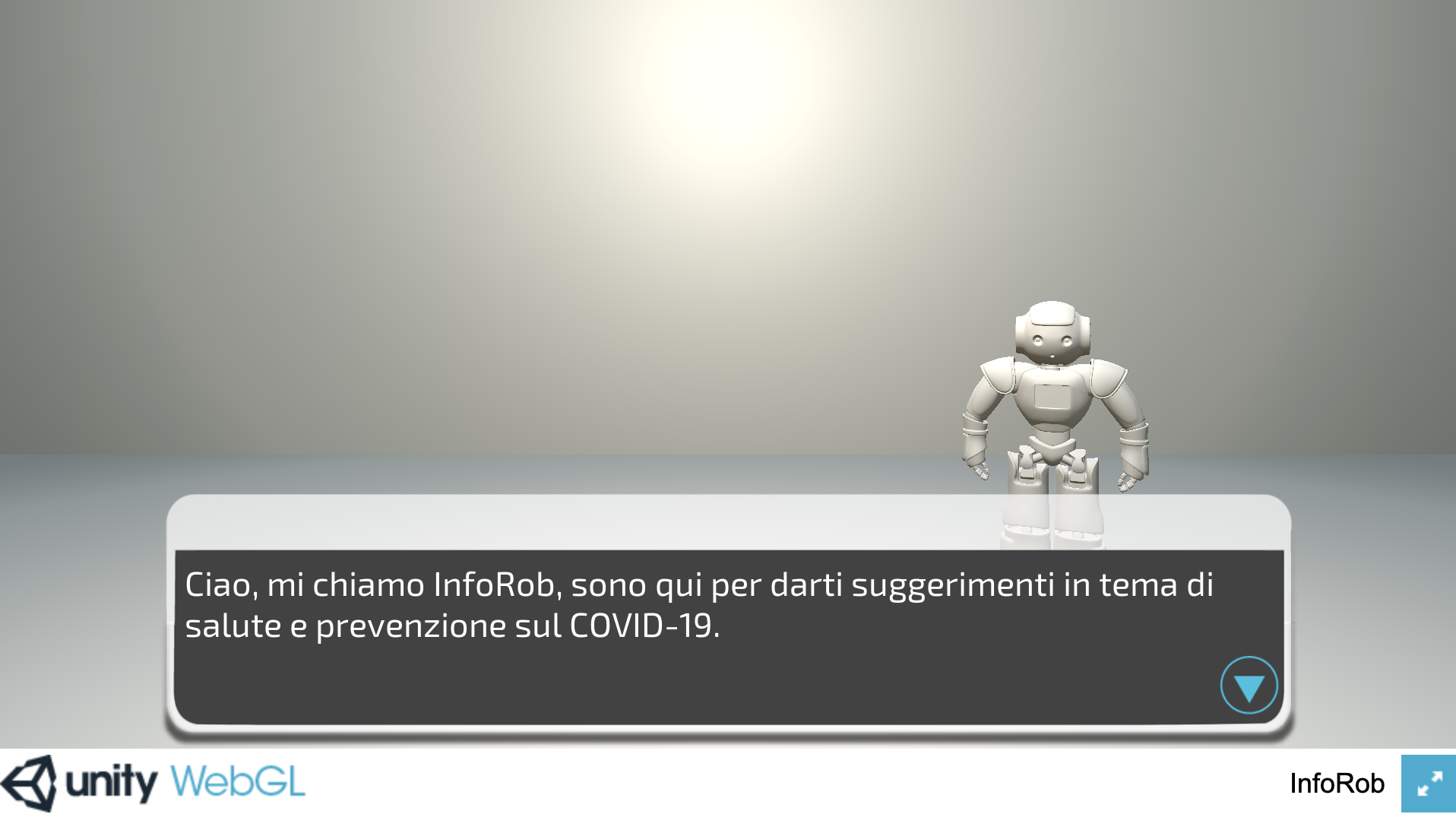}
\end{center}
\caption{A screenshot of the InfoRob system where the robot salutes the users by saying: ``Hello, my name is InfoRob, I am here to give you suggestions concerning health and prevention issues on the topic of COVID-19". The behaviour of the simulated robot is controlled via the ACT-R model presented above. The virtual realization of the robot has been done by connecting ACT-R and the Unity 3D engine.}
\label{gui}
\end{figure}

The ACT-R model has been implemented (and is ready for) for being used in a real robotic platform. However, as mentioned, due to the impossibility of conducting experiments in our laboratory due to the current Covid-19 restrictions, we defined  a virtual version of the robot. 
The system, that we named InfoRob, is accessible  through the GUI interface shown in Figure \ref{gui}).
The system integrates the ACT-R cognitive architecture and the Unity3d engine. To this end, we implemented a communication middleware through a WebSocket, which is responsible for starting and monitoring the conversation and managing the logging.
The WebSocket has been realized in Java language and exploits a Java porting of ACT-R's python interface.
The ACT-R model used for this work is available at 
\url{https://github.com/manuelgentile/inforob}. The deployed web version of the  model is, on the other hand, available at \url{ http://cogsgs.pa.itd.cnr.it/inforob/}.

\begin{figure}[ht!]
\begin{center}
\includegraphics[width=13cm,page=11\textwidth]{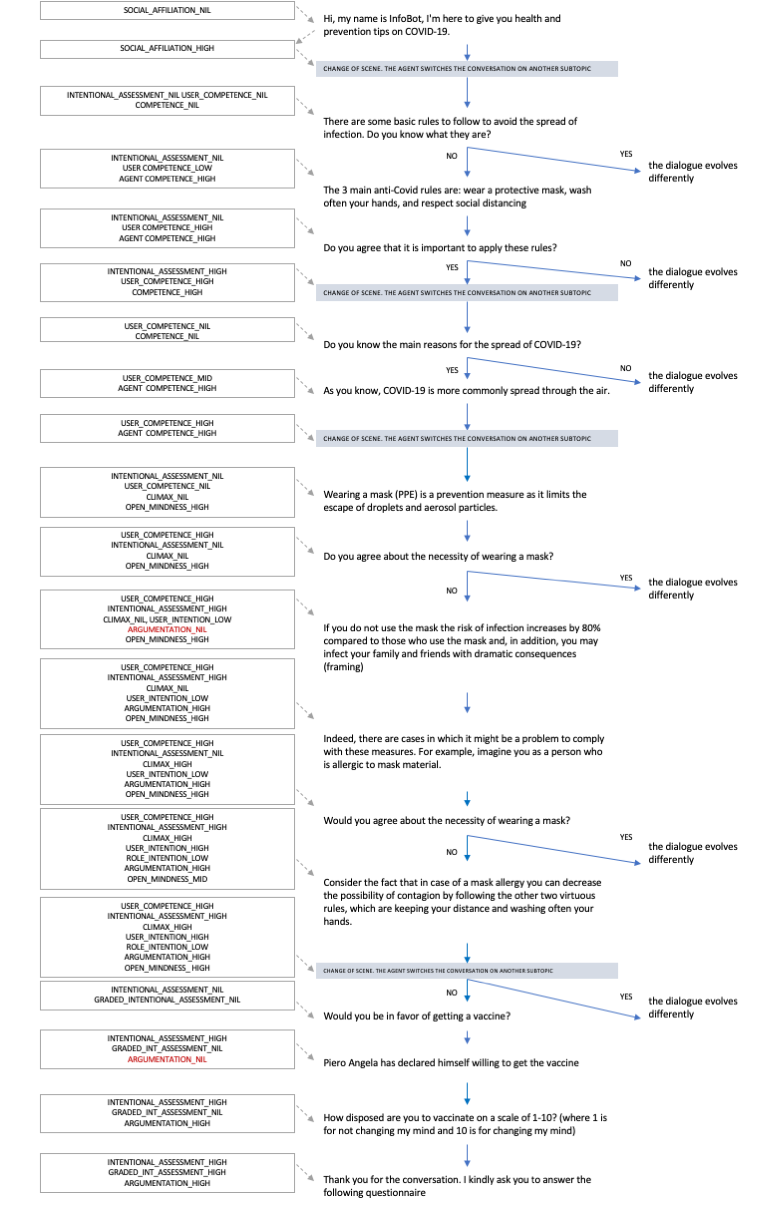} \end{center}
\caption{An example of dialogue evolution reporting the current needs in the main states. The sentences have been translated from the Italian which was the language used by the bot for the dialogue.}
\label{percorso}
\end{figure}

In the following, we provide a simple running example showing all the elements interacting in the ACT-R model in order to determine its own strategies (see Figure \ref{percorso}). 
Starting from the typical topics of discussion in a conversation about COVID-19 rules, we considered the deepening of one topic as possible conversation scenes (e.g. contagion, use of the mask, hands washing, social distancing). Other scenes that are considered are an Introduction and a Conclusion Scene. 
The defined ACT-R model is based on emerging needs in relation to the conversation state. This state of affairs leads the agent to follow a reasoning flow as shown in Figure \ref{flow}. 

The dialogue evolves according to the choices of the user and the ethical profile of the agent. 
Figure \ref{percorso}, as anticipated, shows in detail an overall example of a dialogue evolving  from  a set of unsatisfied needs  emerging during the interaction and the consequent behaviour of the agent. In relation to its needs the agent will select a dialogue act as discussed in section \ref{cogmod}.

As an example, let us consider the 12th dialogue state of the interaction depicted in figure \ref{percorso}: here it is shown  how emerges an Argumentation need since, in the previous passages, the user intention in wearing a mask still remained low despite the information provided by the agent. The emergence of such a need leads the agent to produce an "argument act". In the specific case, the agent randomly chooses a \emph{framing} argumentation  producing the following sentence ``If you do not use the mask, the risk of infection increases by 80\% compared to those who use the mask and, in addition, you may infect your family and friends with dramatic consequences''.

In the 13th state of the dialogue, emerges the need of the agent to reach a climax point, so the agent produces an "exception" act to introduce a situation where a particular condition of the user (a strong allergy) is in conflict with the  rule following previously argued. The sentence associated to the "exception" act is the following: "Indeed there are cases in which it might be a problem to comply with these measures. For example, imagine you as a person who is allergic to mask material".

By following this example, in the 15th state (that follows the 14th one aiming at re-assessing the willingness to follow the mask rule by the user also in its the newly assigned narrative role) even in the eventuality in which the user intention on wearing a mask is low, if the agent is equipped with  an open-mindness ethical profile (a choice that is done randomly at the beginning of each dialogue), it will select a "substitution" act to propose to the user an alternative to manage the conflicting situation generated by the employed storytelling strategy. The sentence associated with a substitution act is "Consider the fact that in case of a mask allergy you can decrease the possibility of contagion by following the other two virtuous rules, which are keeping your distance and washing often your hands."
The dialogue will unfold in this way, i.e. by following a needs-driven model and by employing different persuasive techniques based on the user responses, until the end.

\section{Exploratory Evaluation}
\label{results}

We evaluated the proposed ACT-R model by using the deployed online platform presented before. The selection of the users has been done on an availability sampling strategy. This is a sampling of convenience, based on subjects available to the researcher/team of researchers, often used in small scale experiments when the population source is not completely defined. Even though random sampling is the best way of having a representative sample, these strategies require a great deal of time and money and are less used in qualitative analyses or preliminary quantitative analyses where the correct target population to reach is not easily accessible within a limited time-frame. Therefore, much research in both social psychology and, consequently, in human-computer interaction and human-robot interaction is based on samples obtained through non-random selection [\cite{Royce:1999}], and our work makes no exceptions here. 

In total, 63 people (23 men and 40 women) with a mean age of 39.65 years ($sd=12.04$) participated on a voluntary basis in the trial. All of them were native Italian speakers and all were naive to the experimental procedure and to the aims of the study (they were told that the goal of trial was to evaluate the the level of information about COVID-19-related issues provided by the simulated robot) \footnote{In order to reduce the sampling bias due to our group-networks (i.e. academics and researchers which are likely to be more open minded on the COVID-19 topic compared to the general population) we mainly recruited the participants from open Facebook groups consisting of a large number of people with heterogeneous positions towards anti-COVID19 rules and vaccines (in most cases we explicitly contacted those expressing negative opinions about the vaccines and, after a number of interactions, we were able to convince some of them to participate in our experiment. Another source of recruitment was WhatsApp by exploiting our personal network of contacts. In particular, we asked our contacts to disseminate the link among their contacts via the WhatsApp groups they were part of.  As mentioned above, all the participants were obviously naive to the experiment and to our research goal.}.

At the end of the dialogue, the users were asked to fill a questionnaire, whose translation in English is attached in the APPENDIX below (the original questionnaire was in Italian since the entire dialogue based conversation was also in Italian). All the users were explicitly advised that all the collected data were completely anonymous. The questionnaire was intended to assess three main elements: 1) whether and to what extent the adoption of persuasive and argumentative techniques in the dialogue (such arguments were presented only to the group of users showing no intention of following the rules or of taking a vaccine) can influence the level of agreement expressed by such group on pro-vaccine and pro-COVID-19 rules sentences (please note that the content of these sentences was exactly the one presented to such users in the form of framing or ad populum or ad verecundiam arguments).
2) to what extent the use of a narrative strategy, based on a classical conflict-induced storytelling activity, can influence the re-evaluation by the users of prior beliefs and pre-conceptions about COVID-19 rules and vaccines. 
3) to what extent the use of an ethical stance (based on the Virtue Ethics applied to a dialogical condition, as indicated in \ref{virtue}) influence the overall user perception of the efficacy of the dialogical interaction with a technological artefact.
Overall, the following figure emerged: all the participants except three declared themselves aware of the methods of spreading the virus, and only two of the 63 declared not to know the rules of social distancing, the obligation to wear the mask and washing hands. Five participants expressed opposition to the implementation of these rules.
Regarding the need to wear the mask and to take the vaccine, the system, after evaluating the predisposition of the users towards these two behaviours, in the presence of a low predisposition, proposed to them, as mentioned, an argumentative dialogue act according to the different modalities described above (i.e. populum, verecundiam and framing).

Only four people out of 63 stated that they disagreed with wearing the mask. On the other hand, in the case of whether or not to get a vaccine, 7 out of 63 people said they disagreed.
In this regard, it should be noted that, in general, it is difficult to reach skeptical users willing to explicitly express their negative predisposition towards these topics in this type of  evaluation (exceptions can be be made for people explicitly declaring themselves no-vax that, however, represents a minority of people potentially having skeptical reserves towards anti-COVID-19 vaccines or rules). Typically, these groups tend not to be interviewed fearing some strange use of the data (despite explicitly informed of the ethical and research-wise measures adopted) and - when interviewed -  are not in favour of listening or accommodating arguments contrary to their preconceptions.

The limited collection of this kind of data makes the data analysis partial. However, even from these few interactions, some interesting points can be made.
First, all participants were satisfied by the interaction with the robot-avatar ($M=8.79$,$sd=1.66$).
Concerning the effectiveness of the persuasive arguments used during the interaction, the low numbers of 'contrary users' detected (we remind that only this subgroup of users was the one exposed to persuasive techniques) do not allow for definitive conclusions to be drawn. 
However, from the analysis of Figure 5, it can be seen that in two cases out of three (i.e. in the case of framing and ad verecundiam) the use of such techniques has led - in the questionnaire filled at the end of the interaction -  to an increased level of agreement of such users (with respect to those in the subgroup not exposed to such techniques) with the corresponding statements that were presented to them during the dialogue by making use of such argumentative techniques. In our view, this is a symptom of the fact that, even in this worst-case scenario represented by the difficulty of changing the attitudes of such subjects on topics related to COVID-19 rules and vaccines, such techniques have a persuasive effect (the case is particularly relevant for the ad verecundiam as shown in Figure 5. On the other hand, the argumentum ad populum has sorted an effect contrary to the persuasive intention on the robot-avatar. As we will see in the Discussion section, this datum is compliant with other findings.

 \begin{figure}[h!]
 \label{fig:argumentation_plot}
\begin{center}
\includegraphics[origin=c, width=1\textwidth]{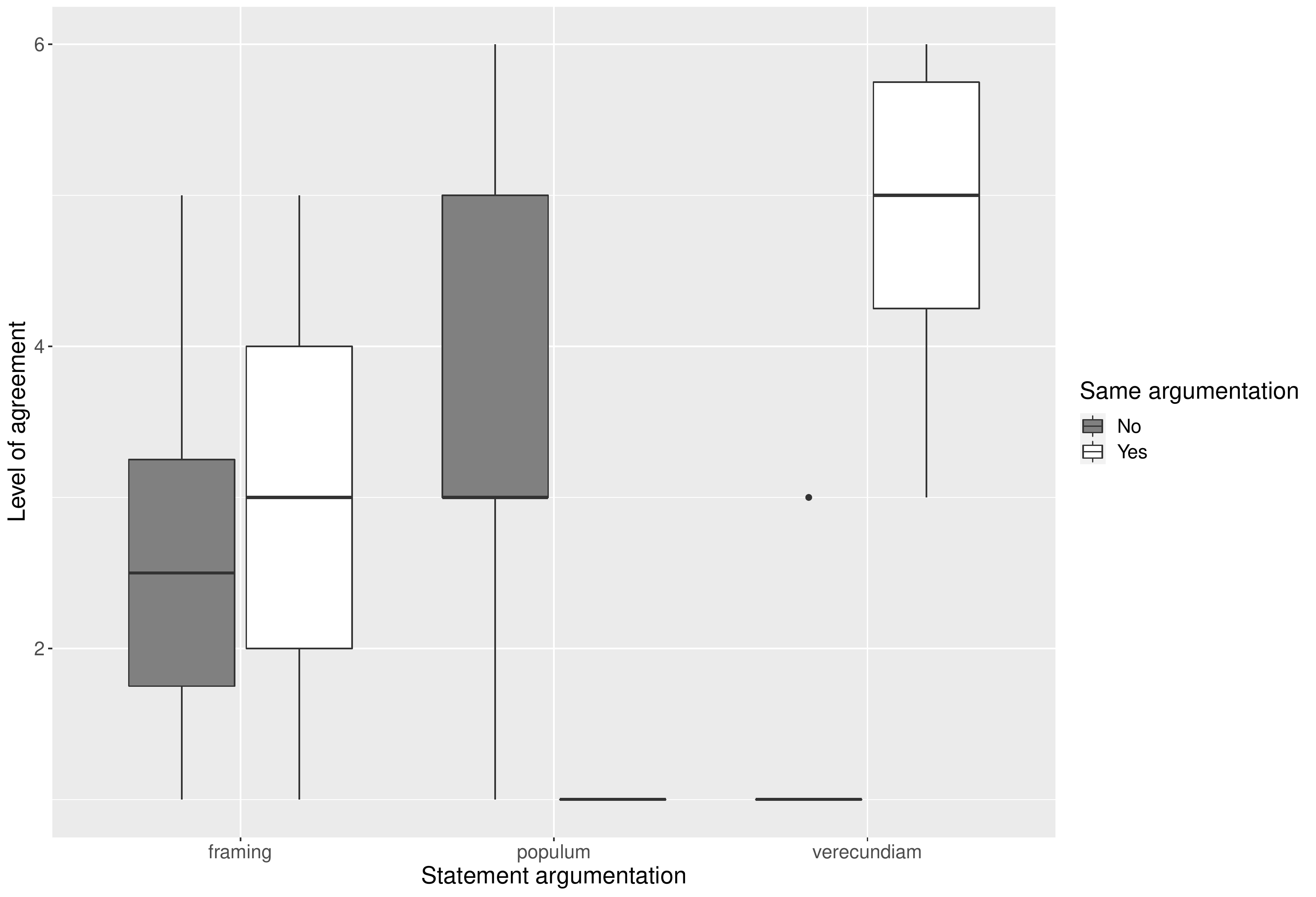} 
\end{center}
\caption{Users' agreements on the statements employed in the persuasive techniques. The agreement is overall major (with the exception of the \emph{ad populum}) in the subjects exposed to such techniques (white bars).}
\end{figure}

Another element supporting the idea that the use of persuasive techniques is effective for improving the attribution (by the same subgroup of users explicitly showing an high degree of distrust on COVID-19 vaccines and rules) of an overall perceived efficacy of the experienced dialogue is shown in Figure 6. 
This figure shows the perceived level of efficacy attributed by such a subgroup of users to the overall dialogue experienced with the robot-avatar. 
Although the data do not show significant differences between the types of argumentative stimuli received and the overall level of assessment provided, the figure shows a sub-optimal but not negative perceived efficacy (i.e. with average votes around 5 and 5.5. on a cardinal scale going from 1 to 10 points). If one considers that this datum is provided by the most skeptic users (i.e. those showing a high degree of distrust on vaccines and anti-COVID 19 rules) the result is quite impressive.

 \begin{figure}[h!]
 \label{fig:efficacy}
\begin{center}
\includegraphics[origin=c, width=1\textwidth]{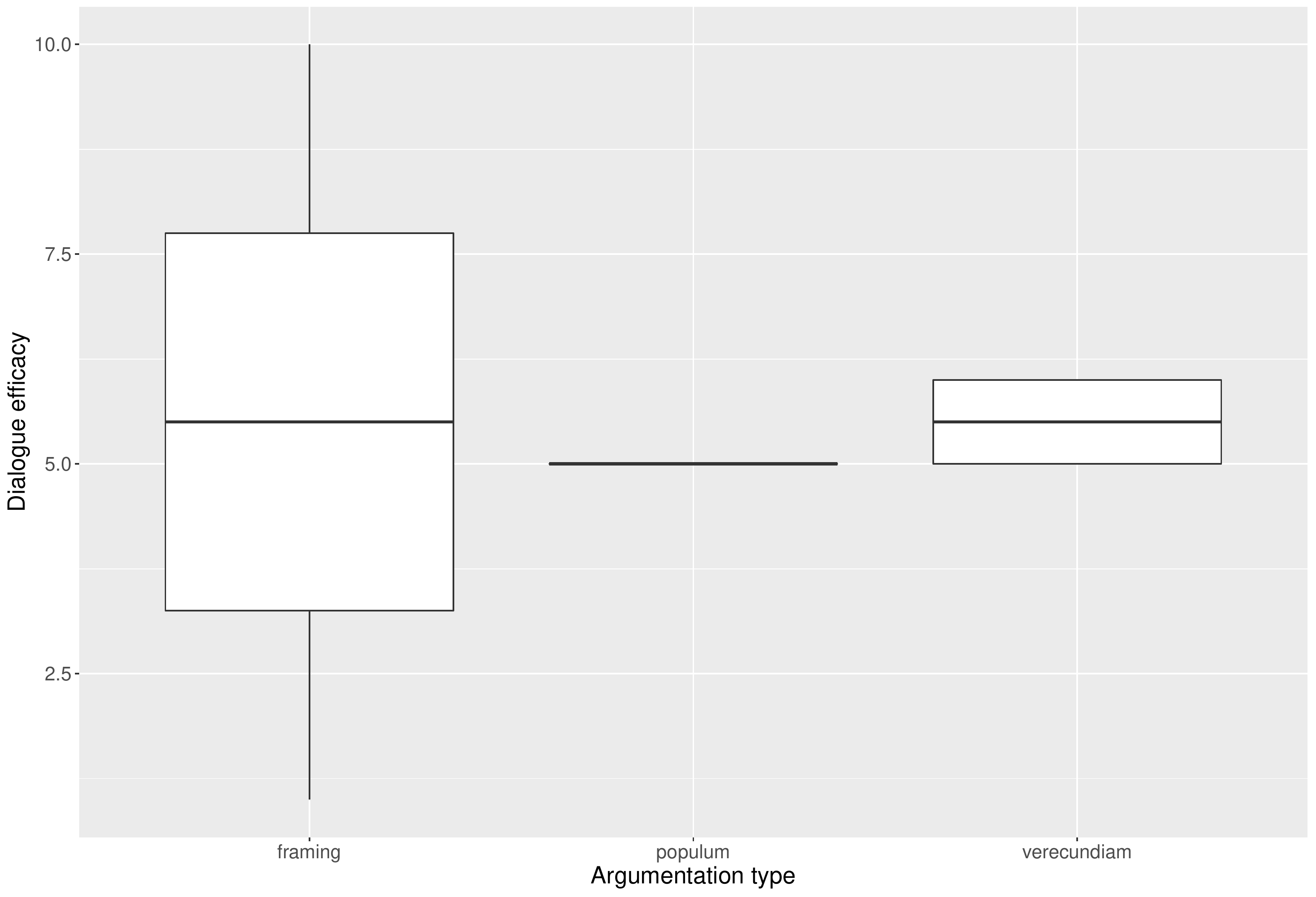} 
\end{center}
\caption{Efficacy of the robot-dialogue rated by the skeptic group of users and dived per argumentation stimulus.}
\end{figure}

Another distinctive element of the dialogue is represented by the evaluation of the storytelling strategy. As mentioned it consisted, for example, in assigning to the users the role of a person who, due to allergic problems, cannot wear any kind of protective mask. Differently from the previous figures, this narrative strategy was tested on all the participants.
In particular, concerning this point, we wanted to verify how much a dialogue guided by the values of open-mindedness, coupled with the use of a persuasive storytelling  strategy, could influence the user's perception and its willingness to eventually re-consider prior beliefs on the assessed topic. 
For this reason, in a completely random way, the system was devised to select (or not) the open-mindedness mode at the beginning of the dialogue.
As can be seen from the Figure 7, the results obtained by users who interacted with the open-minded robot-avatar (i.e. with the avatar whose dialogue unfolding was compliant with the Virtue Ethics, right of the Figure) show  clear effectiveness of the narrative strategy employed by the system in making them re-evaluate their original beliefs when coupled with such an ethical-mode (the average vote is more than 7.5 on a 10 points cardinal scale).

 \begin{figure}[h!]
 \label{fig:openplot}
\begin{center}
\includegraphics[origin=c, width=1\textwidth]{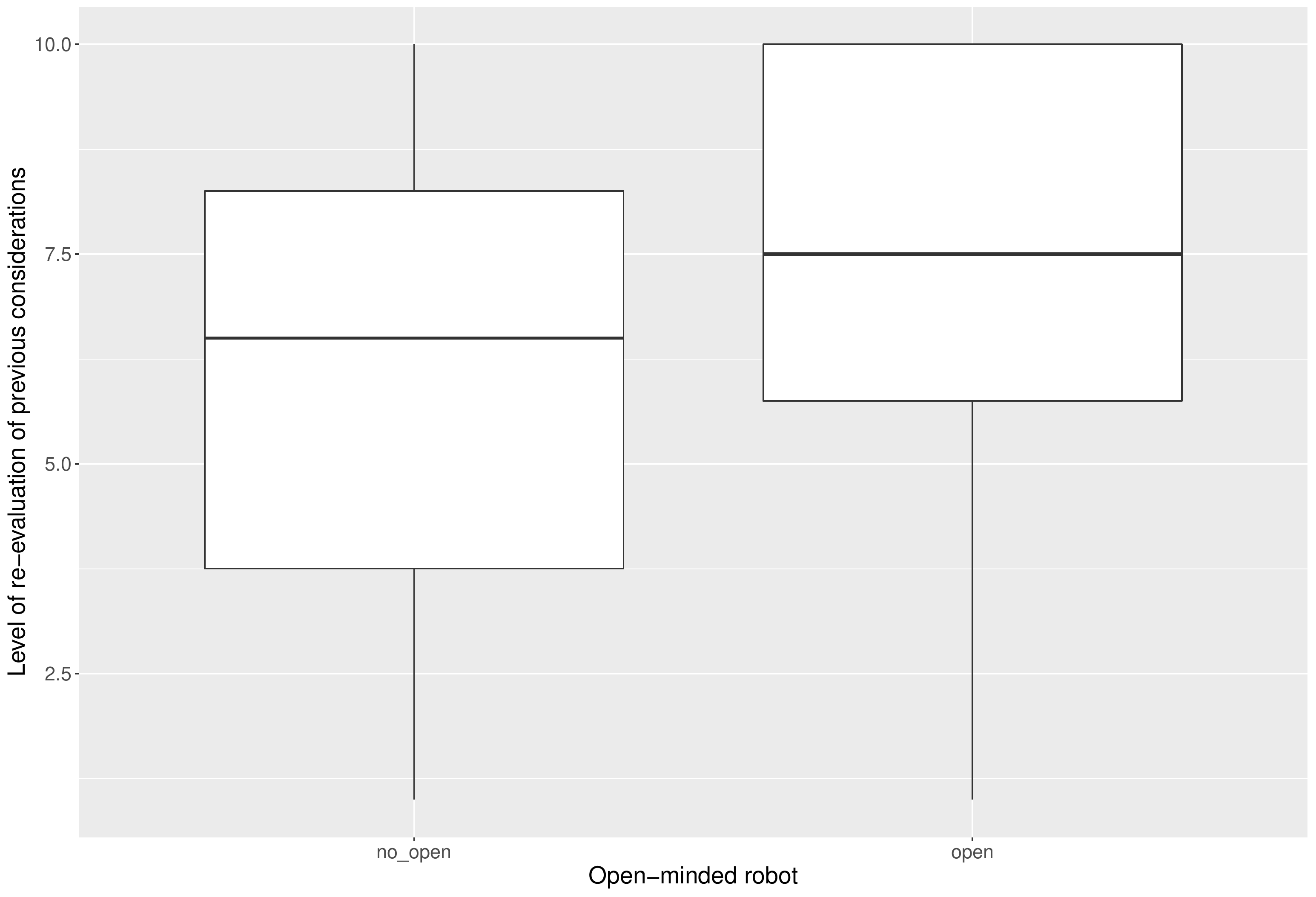} 
\end{center}
\caption{Effect of the ethical dimension (\emph{open-minded} attitude), triggered by the storytelling strategy, in guiding the Users to re-consider their previous beliefs.}
\end{figure}

Finally, we asked the user how much interaction with such a system could help raise awareness of important issues such as this and to what extent interaction with a real robot could be more helpful.
The data show a generally favourable opinion ($M=7.65$ , $sd=2.20$) on the first point, while about the greater effectiveness of a real robot, the users' viewpoint is not so explicit ($M=6.57$ , $sd=2.73$).

\section{Discussions and Future Works}
\label{conclusion}





The dealt topic used for testing our model is quite debated at the moment, and the several argumentations and information  in the main social networks  have led people to consolidate a certain opinion. Moreover, a transition has been observed in the last year from a first phase in which often people did not respect the COVID-19 rules  to a current state in which for example wearing the mask has become, in many cases, a quite consolidated habit. This is reflected in the evaluation, as it is clear that the COVID-19 rules are quite well known and respected. 
The issue regarding vaccines, on the other hand, is more debated, mainly due to the news spread and the short term suspension of one of the vaccines. This has led to more variability in the responses regarding this topic even though, we observed that many people that manifested their distrust on the topic (e.g. in social media or personal conversations) were then against participating in a research experiment that dealt with the COVID topic.

The main figures coming out from such preliminary experimentation, however, points out three elements of interest. First: the use of a storytelling strategy in the dialogue, driven by the assignation of a narrative role to the users, enforces the persuasive strength of the dialogue but only if this dialogue is conducted by following the ethical principles of  virtue ethics. 
Second: the efficacy of the  persuasive techniques coming from classical rhetoric, cognitive psychology and argumentation theory (i.e. argumentum ad populum, ad verecundiam and framing) is not always the same. In particular: the use of ad verecundiam and of the framing techniques seems to provide a persuasive effect if compared to the situation in which these techniques are not used. The ad populum, on the other hand, does
not seem to have any efficacy. As anticipated above, this datum is compliant with other findings exploited in the context of persuasive news and e-commerce item recommendations [\cite{gena2019personalization}]. As a consequence, this technique could be probably left aside in future works since it does not seem to work properly when employed within technological artifacts. 

As already mentioned, the ACT-R equipped simulated robot employed  such persuasive techniques only with the subgroup of users that explicitly indicated, during the dialogue, their opposition to either the  vaccine topic or about the willingness to follow the anti-COVID rules (since those already persuaded of the necessity of these two elements didn't need any additional argument to strengthen their conclusion). 
As a consequence, as expected, the level of agreement showed by such subgroup about the content vehiculated by the adopted persuasive arguments was quite low (despite the emergence of a persuasive effect for the framing and the ad verecundiam technique). As a consequence, it emerges that  using such techniques as sole elements of nudging does not seem sufficient to determine an attitude change with respect to a particular belief on a topic. They can, however, be effectively coupled with other techniques like, for example, the use of a narrative strategy and the adoption of an ethical stance. 

Future works will focus on testing the proposed architecture into a real social robotic platform to exploit a physical robot's embodied features.
Indeed, a real agent's presence can overcome the limitations of the virtual avatar approach in engaging people who tend to be sceptical. The presence of an argumentative robot in a physical environment also allows the monitoring of specific behaviours (such as wearing or not wearing the mask) that can be indicators, more than an explicit question, on the user's attitude and can also directly trigger the persuasive strategy of the robot. In this case, as indicated in the literature review, it will be important to avoid  excessive use of social cues as additional persuasive elements to include in the model to avoid an opposite result.
Moreover, as a short term goal, we are conducting, with the current virtual robots version of the system, a continuation of the experiment in order to increase the sample of participants, the collected data and increase the argumentative richness of the model in terms of topic variety and utterance.

A final remark, also concerning another relevant and interdisciplinary line of investigation that deserves to be detailed in the near future, concerns the assessment of the typology of uses that are considered ethically acceptable for these kinds of techniques in the area of persuasive technologies. We maintain that the adoption of these techniques should entirely fall within the so-called \emph{Ethical Framework for a Good AI Society}[\cite{floridi2018ai4people}].
Therefore, their integration should follow an ``ethics by design" approach, in which ethical and social considerations are taken into account at the design phase of any given system. To this end an element that is somehow taken for granted in the current work concerns the identification, for an artificial agent, of the cases when it is socially legitimate to adopt the kind of techniques outlined in this work. 
More precisely: the current study does not deal with the problem of building a fully autonomous agent that - in principle - should be able to detect and understands in which context it is ethically acceptable to use the above presented persuasive techniques (and in which ones not).
Despite this general objective was out of the scope with respect to the current study, we are aware that a more in-depth analysis of such issue, both from a methodological and experimental point of view, is needed and requires deeper investigations in the near future.

\clearpage\section*{Appendix}
\subsection*{Translated version of the questionnaire}

\begin{itemize}
\item Indicate your gender
\item Please indicate your age
\item How do you rate the level of interaction of the Robot-Avatar (assign a value from 1 to 10 where 1 indicates "poor" and 10 means "excellent")?
\item After interacting with the robot-avatar indicate your level of agreement with the following statements (rate from 1 to 10 where 1 means "not at all agree" and 10 means "completely agree")
\begin{itemize}
\item It is necessary to use the face mask to prevent COVID.
\item If you do not use a mask, the risk of infection increases by 80\% compared to those who use a mask and, in addition, you risk infecting your loved ones with dramatic consequences.
\item As claimed by the vast majority of the population, wearing a mask significantly hinders the virus's spread. 
\item The anti-COVID 19 vaccine is safe, and any side effects are minimal and controllable.   
\item The vaccine is the only solution to exit the health emergency.
\item The vaccine makes it possible not to develop severe forms of the disease (both in its original form and in the English variant).
\end{itemize}
\item Has stepping into the shoes of an allergic person helped you reevaluate your previous considerations? Assign a score from 1 to 10 (1 not at all, 10 completely)
\item Did you get the impression that your interlocutor was willing to take your needs into account during the dialogue? (Assign a score from 1 to 10 to this aspect of the interaction, where 1 indicates "not at all helpful" and 10 indicates "very helpful.")
\item Do you think interacting with such a system would help raise awareness of important issues such as this? (1 not at all helpful - 10 very helpful)
\item Do you think interaction with a real (instead of virtual) robot would help this task? (1 not at all agree - 10 strongly agree)
\end{itemize}

\section*{Compliance with Ethical Standards}

\textbf{Conflict of Interest}: The authors declare that they have no conflict of interest.


%

\bibliographystyle{apalike}

\bibliography{biblio}

\end{document}